\documentclass{sig-alternate}

\usepackage[
pdftitle={Digital Ecosystems: Self-Organisation of Evolving Agent Populations},
pdfauthor={Gerard Briscoe},
pdfsubject={Digital Ecosystems},
pdfkeywords={Digital Ecosystems, evolution, complexity, agent, entropy, clustering, population},
colorlinks=true, linkcolor=black, citecolor=black, filecolor=black, urlcolor=black, hyperfootnotes=false]{hyperref} 
\usepackage[paper=a4paper, asymmetric, left=2cm,right=1.4cm,top=2cm,bottom=3.6cm]{geometry}
\usepackage{graphicx, acronym, ifthen, substr, color}

\usepackage{amsmath}

\newboolean{final}\setboolean{final}{true}

\newboolean{mactex}\setboolean{mactex}{true}

\newboolean{arxiv}\setboolean{arxiv}{false}

\ifthenelse{\boolean{final}}{}{\usepackage{showlabels}\usepackage{citeext}}

\ifthenelse{\boolean{arxiv}}
	{\newcommand{\arxivfootnote}{\footnote}}
	{\def\arxivfootnote#1{}}

\ifthenelse{\boolean{arxiv}}
	{\def\arxivOnly#1{#1}}
	{\def\arxivOnly#1{}}

\ifthenelse{\boolean{arxiv}}
	{\def\arxivNot#1{}}
	{\def\arxivNot#1{#1}}

\ifthenelse{\boolean{arxiv}}
	{}
	{}

\newcommand{\tfigure}[9]
	{
	\IfSubStringInString{!}{#7}{\begin{figure}[#7]}{\begin{figure}[!t]}
	\IfSubStringInString{mm}{#8}{\vspace{#8}}{}
	\centering

	\IfSubStringInString{pdf}{#3}
		{
		\execute{cd images; ln -s #2.pdf .#2.gdf}
		\includegraphics[#1]{images/#2}
		}
		{
		\execute{cd images; ./pdfcrop.sh #2}
		\includegraphics[#1]{images/#2-crop.pdf}
		}

	\vspace{#6}
	\caption[#4]
		{
		\label{#2}
		#4: #5
		}
	\IfSubStringInString{mm}{#9}{\vspace{#9}}{}
	\end{figure}
	}

\newcommand{\Circlesub}[4]
	{
	\ifthenelse{\boolean{mactex}}{}{\immediate\write18{cd images; ./pdfcrop.sh circle#2}}
	\ifthenelse{\boolean{final}}
		{\hspace{#1}\raisebox{#4}{$\includegraphics[scale=0.5, clip=true, trim=0mm 0mm 0mm 0mm]{images/circle#2-crop.pdf}$}\hspace{#3}}
		{\href{file://localhost/Users/g/Desktop/PhDthesis/images/circle#2.graffle}{\hspace{#1}\raisebox{#4}{$\includegraphics[clip=true, trim=0mm 0mm 0mm 0mm]{images/circle#2-crop.pdf}$}\hspace{#3}}}
	}

\newcommand{\squareG}[3]
	{
	\squaresub{#1}{#2}{#3}{-1pt}
	}
	
\newcommand{\squaresub}[4]
	{
	\immediate\write18{cd images; ./pdfcrop.sh square#2}
	\ifthenelse{\boolean{final}}
		{\hspace{#1}\raisebox{#4}{$\includegraphics[scale=0.75,clip=true, trim=0mm 0.25mm 0.25mm 0mm]{images/square#2-crop.pdf}$}\hspace{#3}}
		{\href{file://localhost/Users/g/Desktop/PhDthesis/images/square#2.graffle}{\hspace{#1}\raisebox{#4}{$\includegraphics[scale=0.75,clip=true, trim=0mm 0.25mm 0.25mm 0mm]{images/square#2-crop.pdf}$}\hspace{#3}}}
	}

\newcommand{\execute}[1]{\immediate\write18{#1}}

\definecolor{tred}{RGB}{255,0,0}

\newcommand{\setCap}[2]{#1\immediate\write18{./mkcaption.sh #2}}
\newcommand{\getCap}[1]{\acl*{#1}}

\acrodef{PCG}{Projected Conjugate Gradient} 
\acrodef{QP}{quadratic programming}
\acrodef{RBF}{Radial-Basis Function}
\acrodef{ABM}{Agent-Based Modelling}
\acrodef{AI}{Artificial Intelligence}
\acrodef{DAI}{Distributed Artificial Intelligence}
\acrodef{API}{Application Programming Interface}
\acrodef{ARF}{p14ARF human tumor-suppressor gene}
\acrodef{B2B}{business-to-business}
\acrodef{BDP}{Biological Design Pattern}
\acrodef{BGS}{Best Guess Solution}
\acrodef{BIC}{Biologically-Inspired Computing}
\acrodef{BML}{Business Modelling Language}
\acrodef{BPEL}{Business Process Execution Language}
\acrodef{BPMN}{Business Process Modelling Notation}
\acrodef{CAS}{Complex Adaptive Systems}
\acrodef{COBOL}{COmmon Business-Oriented Language}
\acrodef{DBE}{Digital Business Ecosystem}
\acrodef{DE}{Digital Ecosystem}
\acrodef{DEC}{distributed evolutionary computing}
\acrodef{DGA}{Distributed genetic algorithms}
\acrodef{DIS}{Distributed Intelligence System}
\acrodef{DNA}{Deoxyribose Nucleic Acid}
\acrodef{DOP}{DBE Open Protocol}
\acrodef{DSS}{Distributed Storage System}
\acrodef{EAP}{Evolving Agent Population}
\acrodef{ebXML}{e-business eXtensible Markup Language}
\acrodef{EC}{Evolutionary Computing}
\acrodef{ECJ}{Evolutionary Computing in Java}
\acrodef{EE}{Evolutionary Environment}
\acrodef{EFL}{Evolutionary Framework for Language}
\acrodef{FLE}{Framework for Language Ecosystems}
\acrodef{EOA}{Ecosystem-Oriented Architecture}
\acrodef{ESS}{evolutionary stable strategy}
\acrodef{EvE}{Evolutionary Environment}
\acrodef{ExE}{Execution Environment}
\acrodef{FCB}{Framework for Computational Biomimicry}
\acrodef{FFF}{Fitness Function Framework}
\acrodef{FL}{Fitness Landscape}
\acrodef{HWU}{Heriot-Watt University}
\acrodef{ICL}{Imperial College London}
\acrodef{ICT}{Information and Communications Technology}
\acrodef{INTEL}{Intel Ireland}
\acrodef{IPA}{International Phonetic Alphabet}
\acrodef{ISUFI}{Istituto Superiore Universitario di Formazione Interdisciplinare}
\acrodef{JDJ}{Java Developer's Journal}
\acrodef{KC}{Kolmogorov-Chaitin}
\acrodef{LAN}{local area network}
\acrodef{LSE}{London School of Economics and Political Science}
\acrodef{MAS}{Multi-Agent System}
\acrodef{MDL}{Minimum Description Length}
\acrodef{MDM2}{murine double minute 2}
\acrodef{MFT}{Mean Field Theory}
\acrodef{MoAS}{Mobile Agent System}
\acrodef{MOF}{Meta Object Facility}
\acrodef{MUH}{migration and usage history}
\acrodef{NIC}{Nature Inspired Computing}
\acrodef{NN}{Neural Network}
\acrodef{NoE}{Network of Excellence}
\acrodef{OMG}{Open Mac Grid}
\acrodef{OPAALS}{Open Philosophies for Associative Autopoietic Digital Ecosystems}
\acrodef{P2P}{peer-to-peer}
\acrodef{P53}{protein 53}
\acrodef{PDA}{Personal Digital Assistant}
\acrodef{QoS}{quality of service}
\acrodef{REST}{REpresentational State Transfer}
\acrodef{RNA}{Deoxyribose Nucleic Acid}
\acrodef{SAE}{Software Agent Ecosystem}
\acrodef{SBML}{Systems Biology Modelling Language}
\acrodef{SBVR}{Semantics of Business Vocabulary and Business Rules}
\acrodef{SDL}{Service Description Language}
\acrodef{SF}{Service Factory}
\acrodef{SIM}{Social Interaction Mechanism}
\acrodef{SM}{Service Manifest}
\acrodef{SME}{Small and Medium sized Enterprise}
\acrodef{SML}{Service Modelling Language}
\acrodef{SMO}{Sequential Minimal Optimisation}
\acrodef{SOA}{Service-Oriented Architecture}
\acrodef{SOAP}{Simple Object Access Protocol}
\acrodef{SOC}{Self-Organised Criticality}
\acrodef{SOLUTA}{SOLUTA.NET}
\acrodef{SOM}{Self-Organising Map}
\acrodef{SSL}{Semantic Service Language}
\acrodef{STU}{Salzburg Technical University}
\acrodef{SUN}{Sun Microsystems}
\acrodef{SVM}{Support Vector Machine}
\acrodef{TM}{Turing Machine}
\acrodef{UBHAM}{University of Birmingham}
\acrodef{UDDI}{Universal Description Discovery and Integration}
\acrodef{UML}{Unified Modelling Language}
\acrodef{URI}{Uniform Resource Identifier}
\acrodef{UTM}{Universal Turing Machine}
\acrodef{VLP}{variable length population}
\acrodef{VLS}{variable length sequences}
\acrodef{vls}{variable length sequence}
\acrodef{WP}{Work-Package}
\acrodef{WSDL}{Web Services Definition Language}
\acrodef{XMI}{XML Metadata Interchange}
\acrodef{XML}{eXtensible Markup Language}
\acrodef{MD5}{Message-Digest algorithm 5}
\acrodef{GA}{genetic algorithm}
\acrodef{GP}{genetic programming}
\acrodef{MASON}{Multi-Agent Simulator Of Neighbourhoods}
\acrodef{Repast}{Recursive Porous Agent Simulation Toolkit}
\acrodef{JCLEC}{Java Computing Library for Evolutionary Computing}
\acrodef{OWL-S}{Web Ontology Language - Service}
\acrodef{EGT}{Evolutionary Game Theory}
\acrodef{RBF}{Radial Basis Functions}
\acrodef{SWS}{Semantic Web Services}
\acrodef{HDD}{Hard Disk Drive}
\acrodef{SSD}{Solid-State Drive}
\acrodef{OKS}{Open Knowledge Space}
\acrodef{CAES}{Complex Adaptive EcoSystem}
\acrodef{SaaS}{Software-as-a-Service}
\acrodef{PaaS}{Platform-as-a-Service}
\acrodef{IaaS}{Infrastructure-as-a-Service}
\acrodef{C3}{Community Cloud Computing}

\acrodef{visOrgCap}{agent population on the left intuitively shows organisation through the uniformity of the colours across the agent-sequences, whereas the population to the right shows little}
\acrodef{genCap2}{equalling the maximum length would be incorrect}
\acrodef{orgCPcap}{are consistent with the intuitive understanding one would have for the self-organisation of the sample populations}
\acrodef{largeVisCap}{The visualisation shows that our Efficiency $E$ accurately measures the self-organised complexity of the two populations.}
\acrodef{graph32cap}{The Efficiency tends to a maximum of one, indicating that the population consists of one cluster}
\acrodef{visOrgCap}{agent population on the left intuitively shows organisation through the uniformity of the colours across the agent-sequences, whereas the population to the right shows little}
\acrodef{genCap2}{equalling the maximum length would be incorrect}
\acrodef{orgCPcap}{are consistent with the intuitive understanding one would have for the self-organisation of the sample populations}
\acrodef{largeVisCap}{The visualisation shows that our Efficiency $E$ accurately measures the self-organised complexity of the two populations.}
\acrodef{graph32cap}{The Efficiency tends to a maximum of one, indicating that the population consists of one cluster}
\acrodef{visOrgCap}{agent population on the left intuitively shows organisation through the uniformity of the colours across the agent-sequences, whereas the population to the right shows little}
\acrodef{genCap2}{equalling the maximum length would be incorrect}
\acrodef{orgCPcap}{are consistent with the intuitive understanding one would have for the self-organisation of the sample populations}
\acrodef{largeVisCap}{The visualisation shows that our Efficiency $E$ accurately measures the self-organised complexity of the two populations.}
\acrodef{graph32cap}{The Efficiency tends to a maximum of one, indicating that the population consists of one cluster}
\acrodef{visOrgCap}{agent population on the left intuitively shows organisation through the uniformity of the colours across the agent-sequences, whereas the population to the right shows little}
\acrodef{genCap2}{equalling the maximum length would be incorrect}
\acrodef{orgCPcap}{are consistent with the intuitive understanding one would have for the self-organisation of the sample populations}
\acrodef{largeVisCap}{The visualisation shows that our Efficiency $E$ accurately measures the self-organised complexity of the two populations.}
\acrodef{graph32cap}{The Efficiency tends to a maximum of one, indicating that the population consists of one cluster}
\acrodef{visOrgCap}{agent population on the left intuitively shows organisation through the uniformity of the colours across the agent-sequences, whereas the population to the right shows little}
\acrodef{genCap2}{equalling the maximum length would be incorrect}
\acrodef{orgCPcap}{are consistent with the intuitive understanding one would have for the self-organisation of the sample populations}
\acrodef{largeVisCap}{The visualisation shows that our Efficiency $E$ accurately measures the self-organised complexity of the two populations.}
\acrodef{graph32cap}{The Efficiency tends to a maximum of one, indicating that the population consists of one cluster}
\acrodef{visOrgCap}{agent population on the left intuitively shows organisation through the uniformity of the colours across the agent-sequences, whereas the population to the right shows little}
\acrodef{genCap2}{equalling the maximum length would be incorrect}
\acrodef{orgCPcap}{are consistent with the intuitive understanding one would have for the self-organisation of the sample populations}
\acrodef{largeVisCap}{The visualisation shows that our Efficiency $E$ accurately measures the self-organised complexity of the two populations.}
\acrodef{graph32cap}{The Efficiency tends to a maximum of one, indicating that the population consists of one cluster}
\acrodef{visOrgCap}{agent population on the left intuitively shows organisation through the uniformity of the colours across the agent-sequences, whereas the population to the right shows little}
\acrodef{genCap2}{equalling the maximum length would be incorrect}
\acrodef{orgCPcap}{are consistent with the intuitive understanding one would have for the self-organisation of the sample populations}
\acrodef{largeVisCap}{The visualisation shows that our Efficiency $E$ accurately measures the self-organised complexity of the two populations.}
\acrodef{graph32cap}{The Efficiency tends to a maximum of one, indicating that the population consists of one cluster}
\acrodef{visOrgCap}{agent population on the left intuitively shows organisation through the uniformity of the colours across the agent-sequences, whereas the population to the right shows little}
\acrodef{genCap2}{equalling the maximum length would be incorrect}
\acrodef{orgCPcap}{are consistent with the intuitive understanding one would have for the self-organisation of the sample populations}
\acrodef{largeVisCap}{The visualisation shows that our Efficiency $E$ accurately measures the self-organised complexity of the two populations.}
\acrodef{graph32cap}{The Efficiency tends to a maximum of one, indicating that the population consists of one cluster}
\acrodef{visOrgCap}{agent population on the left intuitively shows organisation through the uniformity of the colours across the agent-sequences, whereas the population to the right shows little}
\acrodef{genCap2}{equalling the maximum length would be incorrect}
\acrodef{orgCPcap}{are consistent with the intuitive understanding one would have for the self-organisation of the sample populations}
\acrodef{largeVisCap}{The visualisation shows that our Efficiency $E$ accurately measures the self-organised complexity of the two populations.}
\acrodef{graph32cap}{The Efficiency tends to a maximum of one, indicating that the population consists of one cluster}
\acrodef{visOrgCap}{agent population on the left intuitively shows organisation through the uniformity of the colours across the agent-sequences, whereas the population to the right shows little}
\acrodef{genCap2}{equalling the maximum length would be incorrect}
\acrodef{orgCPcap}{are consistent with the intuitive understanding one would have for the self-organisation of the sample populations}
\acrodef{largeVisCap}{The visualisation shows that our Efficiency $E$ accurately measures the self-organised complexity of the two populations.}
\acrodef{graph32cap}{The Efficiency tends to a maximum of one, indicating that the population consists of one cluster}
\acrodef{visOrgCap}{agent population on the left intuitively shows organisation through the uniformity of the colours across the agent-sequences, whereas the population to the right shows little}
\acrodef{genCap2}{equalling the maximum length would be incorrect}
\acrodef{orgCPcap}{are consistent with the intuitive understanding one would have for the self-organisation of the sample populations}
\acrodef{largeVisCap}{The visualisation shows that our Efficiency $E$ accurately measures the self-organised complexity of the two populations.}
\acrodef{graph32cap}{The Efficiency tends to a maximum of one, indicating that the population consists of one cluster}
\acrodef{visOrgCap}{agent population on the left intuitively shows organisation through the uniformity of the colours across the agent-sequences, whereas the population to the right shows little}
\acrodef{genCap2}{equalling the maximum length would be incorrect}
\acrodef{orgCPcap}{are consistent with the intuitive understanding one would have for the self-organisation of the sample populations}
\acrodef{largeVisCap}{The visualisation shows that our Efficiency $E$ accurately measures the self-organised complexity of the two populations.}
\acrodef{graph32cap}{The Efficiency tends to a maximum of one, indicating that the population consists of one cluster}
\acrodef{visOrgCap}{agent population on the left intuitively shows organisation through the uniformity of the colours across the agent-sequences, whereas the population to the right shows little}
\acrodef{genCap2}{equalling the maximum length would be incorrect}
\acrodef{orgCPcap}{are consistent with the intuitive understanding one would have for the self-organisation of the sample populations}
\acrodef{largeVisCap}{The visualisation shows that our Efficiency $E$ accurately measures the self-organised complexity of the two populations.}
\acrodef{graph32cap}{The Efficiency tends to a maximum of one, indicating that the population consists of one cluster}
\acrodef{visOrgCap}{agent population on the left intuitively shows organisation through the uniformity of the colours across the agent-sequences, whereas the population to the right shows little}
\acrodef{genCap2}{equalling the maximum length would be incorrect}
\acrodef{orgCPcap}{are consistent with the intuitive understanding one would have for the self-organisation of the sample populations}
\acrodef{largeVisCap}{The visualisation shows that our Efficiency $E$ accurately measures the self-organised complexity of the two populations.}
\acrodef{graph32cap}{The Efficiency tends to a maximum of one, indicating that the population consists of one cluster}

\newcommand{\be}{\begin{equation}}
\newcommand{\eeq}[1]{\label{#1}\end{equation}}

\begin{document}
\CopyrightYear{2008} 
\crdata{978-1-60558-829-2/08/0003}  

\title{Digital Ecosystems:\\Self-Organisation of Evolving Agent Populations}

\numberofauthors{2} 
\author{
\alignauthor
Gerard Briscoe\\
	\affaddr{Digital Ecosystems Lab}\\
	\affaddr{Department of Media and Communications}\\
	\affaddr{London School of Economics}\\
	\affaddr{United Kingdom}\\
	\email{g.briscoe@lse.ac.uk}
\alignauthor
Philippe De Wilde\\
	\affaddr{Intelligent Systems Lab}\\
	\affaddr{Department of Computer Science}\\
	\affaddr{Heriot Watt University}\\
	\affaddr{United Kingdom}\\
	\email{pdw@hw.ac.uk}
}

\maketitle
\begin{abstract}
A primary motivation for our research in Digital Ecosystems is the desire to exploit the self-organising properties of biological ecosystems. Ecosystems are thought to be robust, scalable architectures that can automatically solve complex, dynamic problems. Self-organisation is perhaps one of the most desirable features in the systems that we engineer, and it is important for us to be able to measure self-organising behaviour. We investigate the self-organising aspects of Digital Ecosystems, created through the application of evolutionary computing to \acp{MAS}, aiming to determine a macroscopic variable to characterise the self-organisation of the evolving agent populations within. We study a measure for the self-organisation called Physical Complexity; based on statistical physics, automata theory, and information theory, providing a measure of information relative to the randomness in an organism's genome, by calculating the entropy in a population. We investigate an extension to include populations of variable length, and then built upon this to construct an efficiency measure to investigate clustering within evolving agent populations. Overall an insight has been achieved into where and how self-organisation occurs in our Digital Ecosystem, and how it can be quantified.
\end{abstract}



\section{Introduction}

Digital Ecosystems are distributed adaptive open socio-technical systems, with properties of self-organisation, scalability and sustainability, inspired by natural ecosystems \cite{thesis}, and are emerging as a novel approach to the catalysis of sustainable regional development driven by \acp{SME}. Digital Ecosystems aim to help local economic actors become active players in globalisation, valorising their local culture and vocations, and enabling them to interact and create value networks at the global level \cite{dini2008bid}. With its technical component being the \emph{digital} counterpart of a biological ecosystem, providing for the evolution of software services (agents) in a distributed network \cite{de07oz, dbebkpub}.

We considered the available literature on self-organisation, for its general properties, its application to \acp{MAS} (the dominant technology in Digital Ecosystems), and its application to the evolving agent populations of our Digital Ecosystem. Self-organisation has been around since the late 1940s \cite{ashby}, but has escaped general formalisation despite many attempts \cite{nicolis}. There have instead been many notions and definitions of self-organisation, useful within their different contexts \cite{heylighen2002sso}. They have come from cybernetics \cite{ashby, beer1966dac}, thermodynamics \cite{nicolis}, mathematics \cite{lendaris1964dso}, information theory \cite{shalizi2001cac}, synergetics \cite{haken1977sin}, and other domains \cite{lehn1990psc}. The term \emph{self-organising} is widely used, but there is no generally accepted meaning, as the abundance of definitions would suggest. Therefore, the philosophy of self-organisation is complicated, because organisation has different meanings to different people. So, we would argue that any definition of self-organisation is context dependent, in the same way that a choice of statistical measure is dependent on the data being analysed. Therefore, in the context of our Digital Ecosystems \cite{bionetics, javaOne} we shall further considering its self-organisation.

\section{Self-Organisation}

Proposing a definition for self-organisation faces the \emph{cybernetics} problem of defining \emph{system}, the \emph{cognitive} problem of \emph{perspective}, the \emph{philosophical} problem of defining \emph{self}, and the \emph{context} dependent problem of defining \emph{organisation} \cite{gershenson}. The \emph{system} in this context is an \emph{evolving agent population}, with the replication of individuals from one generation to the next, the recombination of the individuals, and a selection pressure providing a differential fitness between the individuals, which is behaviour common to any evolving population \cite{begon96}. \emph{Perspective} can be defined as the perception of the observer in perceiving the self-organisation of a system \cite{beer1966dac}, matching the intuitive definition of \emph{I will know it when I see it}, which despite making formalisation difficult shows that organisation is \emph{perspective} dependent (i.e. relative to the \emph{context} in which it occurs). In the \emph{context} of an evolutionary system, the observer does not exist in the traditional sense, but is the \emph{selection pressure} imposed by the environment, which \emph{selects} individuals of the population over others based on their \emph{observable} \emph{fitness}. Therefore, consistent with the theoretical biology \cite{begon96}, in an evolutionary system the self-organisation of its population is from the \emph{perspective} of its environment. Whether a system is \emph{self}-organising or being organised depends on whether the process causing the organisation is an internal component of the system under consideration. This intuitively makes sense, and therefore requires one to define the boundaries of the system being considered to determine if the force causing the organisation is internal or external to the system. For an evolving population the force leading to its organisation is the \emph{selection pressure} acting upon it \cite{begon96}, which is formed by the environment of the population's existence and competition between the individuals of the population \cite{begon96}. As these are internal components of an evolving agent population \cite{begon96}, it is a self-organising system.

Now that we have defined, for an evolving agent population, the \emph{system} for which its \emph{organisation} is \emph{context} dependent, the \emph{perspective} to which it is relative, and the \emph{self} by which it is caused, a definition for its \emph{self-organisation} can be considered. The \emph{context}, an evolving agent population in its environment, lacks a 2D or 3D metric space, so it is necessary to consider a visualisation in a more abstract form. We will let a single square, \linebreak \squareG{-4mm}{White}{-2.5mm}, represent an agent, with colours to represent different agents. Agent-sequences will therefore be represented by a sequence of coloured squares, \squareG{-6mm}{RedGreenBlue}{-3mm}, with a population consisting of multiple agent-sequences, as shown in Figure \ref{sampleAgentPopulation}. In Figure \ref{sampleAgentPopulation} the number of agents, in total and of each colour, is the same in both populations. However, the \setCap{agent population on the left intuitively shows organisation through the uniformity of the colours across the agent-sequences, whereas the population to the right shows little}{visOrgCap} or no organisation.

\tfigure{width=3.3in}{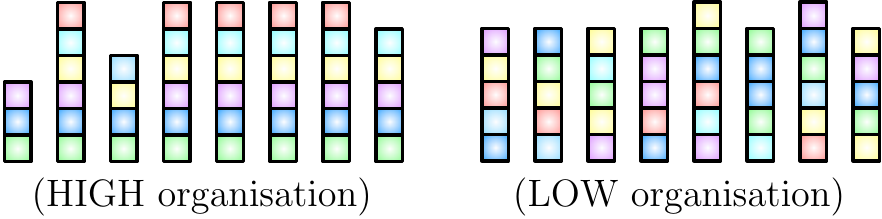}{graffle}{Self-Organisation in Evolving Agent Populations}{The \getCap{visOrgCap} organisation.}{-7mm}{}{}{-4mm}

While alternative definitions have been proposed \cite{crutchfield2006, barron, prugel, chaitin}, with each defining what property or properties demonstrate self-organisation, they lack applicability to evolving agent populations, because of the context dependent nature of self-organisation. However, the properties of Physical Complexity \cite{adami20002} closely matched our intuitive understanding, and so was chosen for further investigation.

\subsection{Physical Complexity}
\label{measureSelfOrg}

Physical Complexity was born \cite{adami1998ial} from the need to determine the proportion of information in sequences of DNA, because it has long been established that the information contained is not directly proportional to the length, known as the C-value enigma/paradox \cite{thomasjr1971goc}. However, because Physical Complexity analyses an ensemble of DNA sequences, the consistency between the different solutions shows the information, and the differences the redundancy \cite{adami2003}. Entropy, a measure of disorder, is used to determine the redundancy from the information in the ensemble. Physical Complexity therefore provides a context-relative definition for the self-organisation of a population without needing to define the context (environment) explicitly \cite{adami2000}.

\label{defPhyCom}
Physical Complexity was derived \cite{adami2000} from the notion of \emph{conditional complexity} defined by Kolmogorov, which is different from traditional Kolmogorov complexity and states that the determination of complexity of a sequence is conditional on the environment in which the sequence is interpreted \cite{li1997ikc}. So, the complexity of a population $S$, of sequences $s$,
\begin{equation}
C = \ell - \sum\limits_{i = 1}^\ell {H(i)}, 
\label{complexity}
\end{equation}
is the maximal entropy of the population (equivalent to the length of the sequences) $\ell$, minus the sum, over the length $\ell$, of the per-site entropies $H(i)$,
\begin{equation}
H(i) = - \sum\limits_{d \in D} {p_d (i)\log _{|D|} p_d (i)}, \\
\label{persite}
\end{equation}
where $i$ is a site in the sequences ranging between one and the length of the sequences $\ell$, $D$ is the alphabet of characters found in the sequences, and $p_d(i)$ is the probability that site $i$ (in the sequences) takes on character $d$ from the alphabet $D$, with the sum of the $p_d(i)$ probabilities for each site $i$ equalling one, $\sum\limits_{d \in D} {p_d (i) = 1}$ \cite{adami2000}. So, the equivalence of the maximum complexity to the length matches the intuitive understanding that if a population of sequences of length $\ell$ has no redundancy, then their complexity is their length $\ell$.

If $G$ represents the set of all possible genotypes constructed from an alphabet $D$ that are of length $\ell$, then the size (cardinality) of $|G|$ is equal to the size of the alphabet $|D|$ raised to the length $\ell$,
\begin{equation}
|G| = |D|^\ell.
\label{recPopSize}
\end{equation}
For the complexity measure to be accurate, a	sample size of $|D|^\ell$ is suggested to minimise the error \cite{adami2000}, but such a large quantity can be computationally infeasible. The definition's creator, for practical applications, chooses a population size of $|D|\ell $, which is sufficient to show any trends present. So, for a population of sequences $S$ we choose, with the definition's creator, a computationally feasible population size of $|D|$ times $\ell $,
\begin{equation}
|S|\ \ge \ |D|\ell.
\label{popSize}
\end{equation}
The size of the alphabet, $|D|$, depends on the domain to which Physical Complexity is applied. For RNA the alphabet is the four nucleotides, $D = \{A, C, G, U\}$, and therefore $|D|=4$ \cite{adami2000}.

\subsection{Variable Length Sequences}

Physical Complexity is currently formulated for a population of sequences of the same length \cite{adami2000}, and so we will now investigate an extension to include populations of \aclp{vls}, which will include populations of variable length agent-sequences of our Digital Ecosystem. This will require changing and re-justifying the fundamental assumptions, specifically the conditions and limits upon which Physical Complexity operates. In (\ref{complexity}) the Physical Complexity, $C$, is defined for a population of sequences of length $\ell$ \cite{adami2000}. The most important question is what does the length $\ell$ equal if the population of sequences is of variable length? The issue is what $\ell$ represents, which is the maximum possible complexity for the population \cite{adami2000}, which we will call the \emph{complexity potential} $C_P$. The maximum complexity in (\ref{complexity}) occurs when the per-site entropies sum to zero, $\sum\limits_{i = 1}^\ell {H(i)} \to 0$, as there is no randomness in the sites (all contain information), i.e. $C \to \ell$ \cite{adami2000}. So, the \emph{complexity potential} equals the length,
\begin{equation}
C_P = \ell,
\label{comPot2} 
\end{equation}
provided the population $S$ is of sufficient size for accurate calculations, as found in (\ref{popSize}), i.e. $|S|$ is equal or greater than $|D|\ell$. For a population of \aclp{vls}, $S_{V}$, the complexity potential, $C_{V_P}$, cannot be equivalent to the length $\ell$, because it does not exist. However, given the concept of minimum sample size from (\ref{popSize}), there is a length for a population of \aclp{vls}, $\ell_V$, between the minimum and maximum length, such that the number of per-site samples up to and including $\ell_V$ is sufficient for the per-site entropies to be calculated. So the \emph{complexity potential} for a population of \aclp{vls}, $C_{V_P}$, will be equivalent to its \emph{calculable} length, 
\begin{equation}
\label{potential}
C_{V_P} = \ell_V.
\end{equation}

If $\ell_V$ where to be equal to the length of the longest individual(s) $\ell _{max}$ in a population of \aclp{vls} $S_{V}$, then the operational problem is that for some of the later sites, between one and $ \ell _{max}$, the sample size will be less than the population size $|S_{V}|$. So, having the length $\ell_V$ \setCap{equalling the maximum length would be incorrect}{genCap2}, as there would be an insufficient number of samples at the later sites, and therefore $\ell _V \not\equiv \ell _{max}$. So, the length for a population of \aclp{vls}, $\ell _V $, is the highest value within the range of the minimum (one) and maximum length, $1 \le \ell _V \le \ell _{\max } $, for which there are sufficient samples to calculate the entropy. A function which provides the sample size at a given site is required to specify the value of $\ell_V$ precisely,
\begin{equation}
sampleSize(i\ :site)\ :int,
\end{equation}
where the output varies between $1$ and the population size $|S_{V}|$ (inclusive). Therefore, the length of a population of \aclp{vls}, $\ell_{V}$, is the highest value within the range of one and the maximum length for which the sample size is greater than or equal to the alphabet size multiplied by the length $\ell _{V}$,
\begin{equation}
sampleSize(\ell _V ) \ge |D|\ell _V \wedge sampleSize(\ell _V + 1) < |D|\ell _V,
\label{lengthVLP}
\end{equation}
where $\ell _V$ is the length for a population of \aclp{vls}, and $\ell _{max} $ is the maximum length in a population of \aclp{vls}, $\ell _V$ varies between $ 1 \le \ell _V \le \ell _{max }$, $D$ is the alphabet and $|D| > 0$. This definition intrinsically includes a minimum size for populations of \aclp{vls}, $|D|\ell _V$, and therefore is the counterpart of (\ref{popSize}), which is the minimum population size for populations of fixed length.

The length $\ell$ used in the limits of (\ref{persite}) no longer exists, and therefore (\ref{persite}) must be updated; so, the per-site entropy calculation for \aclp{vls} will be denoted by $H_{V}(i)$, and is, 
\begin{equation}
H_V (i) = - \sum\limits_{d \in D} {p_d (i)\log _{|D|} p_d (i)},
\label{perSiteVLP}
\end{equation}
where $D$ is still the alphabet, $\ell _V$ is the length for a population of \aclp{vls}, with the site $i$ now ranging between $ 1 \le i \le \ell _V $, while the $p_d (i)$ probabilities still range between $ 0 \le p_d (i) \le 1$, and still sum to one. It remains algebraically almost identical to (\ref{persite}), but the conditions and constraints of its use will change, specifically $\ell$ is replaced by $\ell_{V}$. Naturally, $H_{V}(i)$ ranges between zero and one, as did $H(i)$ in (\ref{persite}). So, when the entropy is maximum the character found in the site $i$ is uniformly random, and therefore holds no information.

Therefore, the complexity for a population of \aclp{vls}, $C_{V}$, is the \emph{complexity potential} of the population of \aclp{vls} minus the sum, over the length of the population of \aclp{vls}, of the per-site entropies (\ref{perSiteVLP}),
\begin{equation}
\label{newComplexity}
C_{V} = \ell _V - \sum\limits_{i = 1}^{\ell _V } {H_V (i)},
\end{equation}
where $\ell_V$ is the length for the population of \aclp{vls}, and $H_V(i)$ is the entropy for a site $i$ in the population of \aclp{vls}. 
	
\begin{figure}
	\centering	
	\execute{cd images; ./pdfcrop.sh abstractAgentPopulation}
			\ifthenelse{\boolean{final}}
				{\includegraphics[width=3.3in]{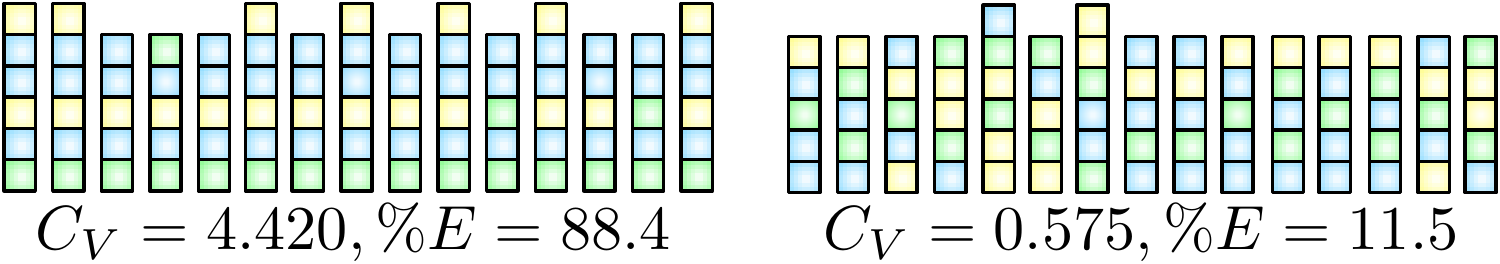}}
				{\href{file://localhost/Users/g/Desktop/PhDthesis/images/abstractAgentPopulation.graffle}{\includegraphics[width=3.33in]{images/abstractAgentPopulation-crop.pdf}}}
	\vspace{-7mm}
	\caption{\label{orgCompPop}Abstract Visualisation for Populations of Variable Length Sequences: The Physical Complexity and Efficiency values \getCap{orgCPcap}.}

	\end{figure}

Physical Complexity can now be applied to populations of \aclp{vls}, so we will consider the abstract example populations in Figure \ref{orgCompPop}. We will let a single square, \squareG{-2mm}{White}{-2mm}, represent a site $i$ in the sequences, with different colours to represent the different values. Therefore, a sequence of sites will be represented by a sequence of coloured squares, \squareG{-2mm}{YellowGreenPurple}{-1.5mm}. Furthermore, the alphabet $D$ is the set \{\squaresub{-1.5mm}{Yellow}{-0.5mm}{-2pt}, \squaresub{-2.25mm}{Green}{-0.75mm}{-2pt}, \squaresub{-2mm}{Purple}{-1mm}{-2pt}\}, the maximum length $\ell _{max}$ is 6 and the  length for populations of \aclp{vls} $\ell _V$ is calculated as 5 from (\ref{lengthVLP}). The Physical Complexity values in Figure \ref{orgCompPop} \setCap{are consistent with the intuitive understanding one would have for the self-organisation of the sample populations}{orgCPcap}; the population with high Physical Complexity has a little randomness, while the population with low Physical Complexity is almost entirely random. 

\vspace{2mm}
\section{Efficiency}

Using our extended Physical Complexity we can construct a measure showing the use of the information space, called the Efficiency $E$, which is calculated by the Physical Complexity $C_{V}$ over the complexity potential $C_{V_P }$,
\begin{equation}
E = \frac{{C_V }}{{C_{V_P } }}.
\label{efficiencyEQ}
\end{equation}
The Efficiency $E$ will range between zero and one, only reaching its maximum when the actual complexity $C_{V}$ equals the complexity potential $C_{V_{P}}$, indicating that there is no randomness in the population. In Figure \ref{orgCompPop} the populations of sequences are shown with their respective Efficiency values as percentages, and the values are as one would expect. 

The complexity $C_{V}$ (\ref{newComplexity}) is an absolute measure, whereas the Efficiency $E$ (\ref{efficiencyEQ}) is a relative measure (based on the complexity $C_{V}$). So, the Efficiency $E$ can be used to compare the self-organised complexity of populations, independent of their size, their length, and whether their lengths are variable or not (as it is equally applicable to the fixed length populations of the original Physical Complexity).

\section{Simulation and Results}

A simulated population of agent-sequences, $[A_1, A_1, A_2, ...]$, was evolved to solve user requests, seeded with agents and agents from the \emph{agent-pool} of the habitats in which they were instantiated. A dynamic population size was used to ensure exploration of the available combinatorial search space, which increased with the average size of the population's agents. The optimal combination of agents (agent) was evolved to the user request $R$, by an artificial \emph{selection pressure} created by a \emph{fitness function} generated from the user request $R$. An individual (agent) of the population consisted of a set of attributes, ${a_1, a_2, ...}$, and a user request consisted of a set of required attributes, ${r_1, r_2, ...}$. So, the \emph{fitness function} for evaluating an individual agent $A$, relative to a user request $R$, was
\begin{equation}
fitness(A,R) = \frac{1}{1 + \sum_{r \in R}{|r-a|}},
\label{ff}
\end{equation}
where $a$ is the member of $A$ such that the difference to the required attribute $r$ was minimised. Equation \ref{ff} was used to assign \emph{fitness} values between 0.0 and 1.0 to each individual of the current generation of the population, directly affecting their ability to replicate into the next generation. The evolutionary computing process was encoded with a low mutation rate, a fixed selection pressure and a non-trapping fitness function (i.e. did not get trapped at local optima). The type of selection used \emph{fitness-proportional} and \emph{non-elitist}, \emph{fitness-proportional means that the \emph{fitter} the individual the higher its probability of} surviving to the next generation. \emph{Non-elitist} means that the best individual from one generation was not guaranteed to survive to the next generation; it had a high probability of surviving into the next generation, but it was not guaranteed as it might have been mutated. \emph{Crossover} (recombination) was then applied to a randomly chosen 10\% of the surviving population, a \emph{one-point crossover}, by aligning two parent individuals and picking a random point along their length, and at that point exchanging their tails to create two offspring. \emph{Mutations} were then applied to a randomly chosen 10\% of the surviving population; \emph{point mutations} were randomly located, consisting of \emph{insertions} (an agent was inserted into an agent-sequence), \emph{replacements} (an agent was replaced in an agent-sequence), and \emph{deletions} (an agent was deleted from an agent-sequence). The issue of bloat was controlled by augmenting the \emph{fitness function} with a \emph{parsimony pressure} which biased the search to shorter agent-sequences, evaluating larger than average agent-sequences with a reduced \emph{fitness}, and thereby providing a dynamic control limit which adapted to the average length of the ever-changing evolving agent populations.

Figure \ref{phycom} shows, for a typical evolving agent population, the Physical Complexity $C_V$ (\ref{newComplexity}) for \aclp{vls} and the \emph{maximum fitness} $F_{max}$ over the generations. It shows that the fitness and our extended Physical Complexity; both increase over the generations, synchronised with one another, until generation 160 when the \emph{maximum fitness} tapers off more slowly than the Physical Complexity. At this point the optimal length for the sequences is reached within the simulation, and so the advent of new fitter sequences (of the same of similar length) creates only minor fluctuations in the Physical Complexity, while having a more significant effect on the \emph{maximum fitness}. The similarity of the graph in Figure \ref{phycom} to the graphs in \cite{adami20002} confirms that the Physical Complexity measure has been successfully extended to \aclp{vls}.

\tfigure{width=3.3in}{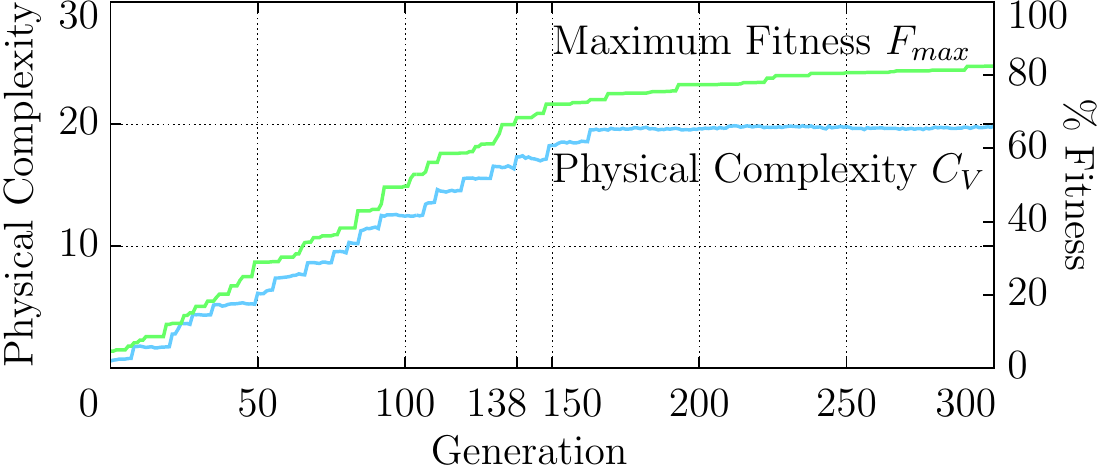}{graph}{Graph of Physical Complexity and Maximum Fitness over the Generations}{The Physical Complexity for \aclp{vls} increases over the generations, showing short-term decreases as expected, such as at generation 138.}{-7mm}{}{}{}

\subsection{Efficiency}

\tfigure{width=3.3in}{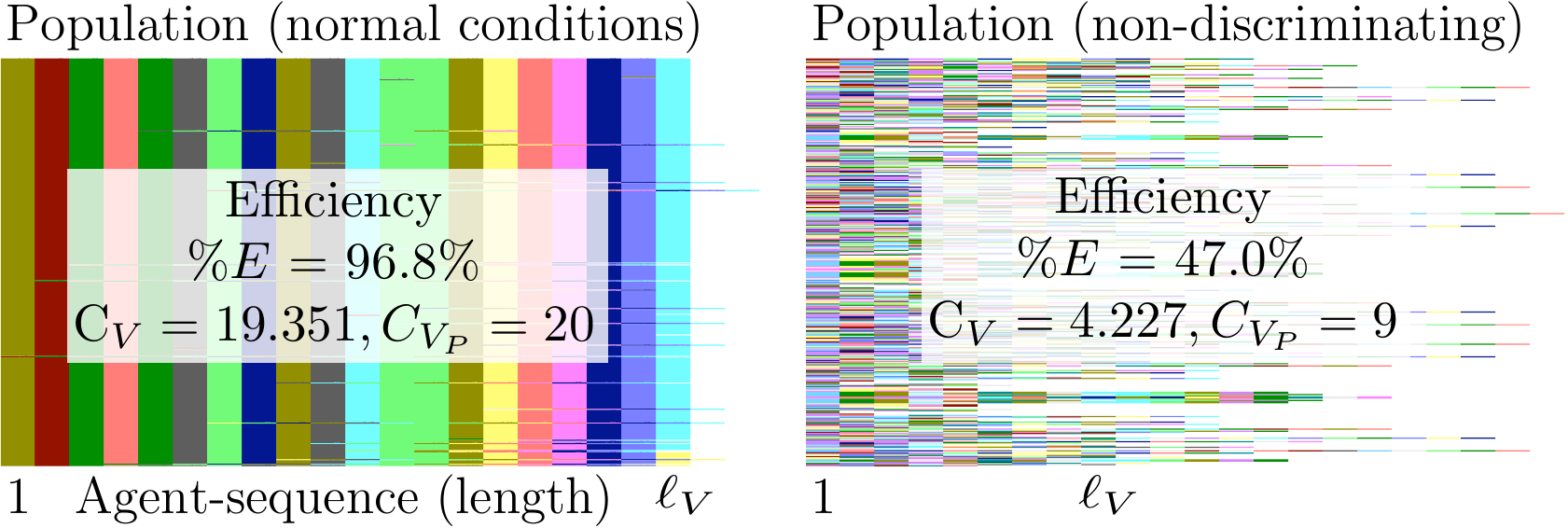}{graffle}{Visualisation of Evolving Agent populations at the 1000th Generation}{The population on the left from Figure \ref{phycom} was run under normal conditions, while the one on the right was run with a non-discriminating selection pressure.}{-7mm}{!b}{}{}

\tfigure{width=3.3in}{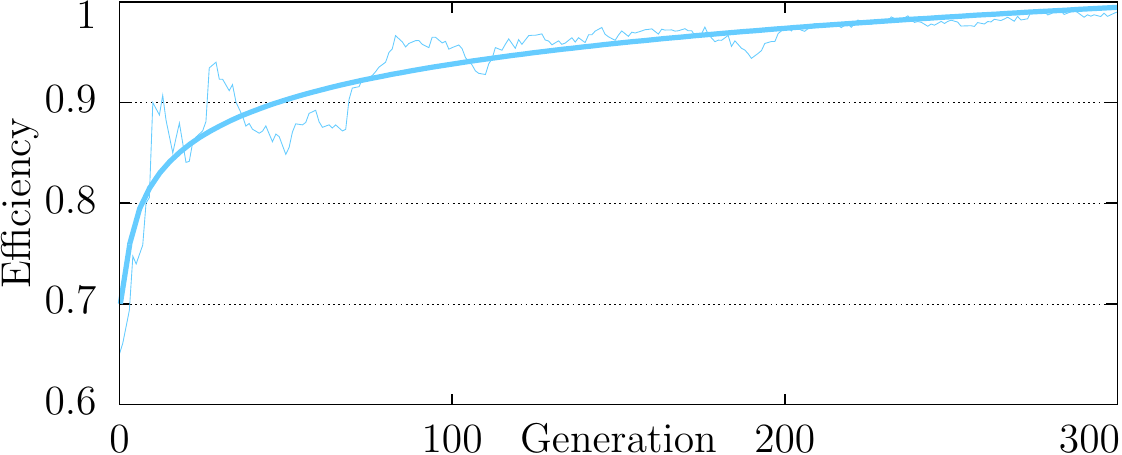}{graph}{Graph of Population Efficiency over the Generations for the population from Figure \ref{phycom}}{\getCap{graph32cap}.}{-7mm}{!b}{}{}

Figure \ref{newphycomvis} is a visualisation of the simulation, showing two alternate populations that were run for a thousand generations, with the one on the left from Figure \ref{phycom} run under normal conditions, while the one on the right was run with a non-discriminating selection pressure. Each multi-coloured line represents an agent-sequence, while each colour represents an agent (site). \setCap{The visualisation shows that our Efficiency $E$ accurately measures the self-organised complexity of the two populations.}{largeVisCap}

Figure \ref{efficiency} shows the Efficiency $E$ (\ref{efficiencyEQ}), over the generations, for the population from Figure \ref{phycom}. \setCap{The Efficiency tends to a maximum of one, indicating that the population consists of one cluster}{graph32cap}, which is confirmed by the visualisation of the population in Figure \ref{newphycomvis} (left).

\section{Conclusions}

We extended Physical Complexity to provide a greater understanding of \acp{MAS} with \emph{evolutionary dynamics}, specifically evolving agent populations, including our \emph{Digital Ecosystem}. We then built upon this to define an \emph{Efficiency} measure. Collectively, the experimental results confirm that Physical Complexity has been successfully extended to evolving agent populations. Most significantly, Physical Complexity has been reformulated algebraically for populations of \aclp{vls}, which we have confirmed experimentally through simulations.

Our \emph{Efficiency} definition not only provides a macroscopic value to characterise the level of \emph{self-organisation}, but the understanding and techniques we have developed have applicability beyond evolving agent populations; as wide as the original Physical Complexity, which has been applied from \acs{DNA} \cite{adami2000} to simulations of self-replicating programmes \cite{adami20032}.

\section{Acknowledgments}

The authors would like to thank the following for encouragement and suggestions; Dr Paolo Dini of the London School of Economics and Political Science, and Dr Christoph Adami of the California Institute of Technology. This work was supported by the EU-funded \ac{OPAALS} Network of Excellence (NoE), Contract No. FP6/IST-034824.

\bibliographystyle{abbrv}
\bibliography{references} 
\end{document}